\documentclass[pdflatex,sn-mathphys-num]{sn-jnl}

\usepackage{graphicx}%
\usepackage{multirow}%
\usepackage{amsmath,amssymb,amsfonts}%
\usepackage{amsthm}%
\usepackage{mathrsfs}%
\usepackage[title]{appendix}%
\usepackage{xcolor}%
\usepackage{textcomp}%
\usepackage{manyfoot}%
\usepackage{booktabs}%
\usepackage{algorithm}%
\usepackage{algorithmicx}%
\usepackage{algpseudocode}%
\usepackage{listings}%
\usepackage{orcidlink}
\definecolor{owncolor}{rgb}{0.95,0.95,0.95}

\theoremstyle{thmstyleone}%
\theoremstyle{thmstyletwo}%

\theoremstyle{thmstylethree}%

\begin{document}

\title[Article Title]{TRAVELER: A Benchmark for Evaluating \textbf{T}emporal \textbf{R}easoning \textbf{a}cross \textbf{V}agu\textbf{e}, Imp\textbf{l}icit and \textbf{E}xplicit \textbf{R}eferences}


\author*[1]{\fnm{Svenja} \sur{Kenneweg}\orcidlink{0009-0002-3025-7563}}\email{skenneweg@techfak.uni-bielefeld.de}
\author[2]{\fnm{Jörg} \sur{Deigmöller}\orcidlink{0009-0007-5931-6973}}\email{Joerg.Deigmoeller@honda-ri.de}
\author[1]{\fnm{Philipp} \sur{Cimiano}\orcidlink{0000-0002-4771-441X}}\email{cimiano@techfak.uni-bielefeld.de}
\author[2]{\fnm{Julian} \sur{Eggert}\orcidlink{0000-0003-4437-6133}}\email{ Julian.Eggert@honda-ri.de}

\affil*[1]{\orgname{Bielefeld University}, \orgaddress{\city{Bielefeld}, \country{Germany}}}
\affil*[2]{\orgname{Honda Research Institute Europe}, \orgaddress{\city{Offenbach}, \country{Germany}}}

\abstract{Understanding and resolving temporal references is essential in Natural Language Understanding as we often refer to the past or future in daily communication. Although existing benchmarks address a systems ability to reason about and resolve temporal references, systematic evaluation of specific temporal references remains limited. Towards closing this gap, we introduce \textbf{TRAVELER}, a novel synthetic benchmark dataset that follows a Question Answering paradigm and consists of questions involving temporal references with the corresponding correct answers. \textbf{TRAVELER} assesses models' abilities to resolve \emph{explicit}, \emph{implicit relative to speech time}, and \emph{vague} temporal references. Beyond investigating the performance of state-of-the-art LLMs depending on the type of temporal reference, our benchmark also allows to evaluate the performance in relation to the length of the set of events. For the category of \emph{vague} temporal references, ground‐truth answers were established via human surveys on Prolific, following a procedure similar to the one from \citet{kenneweg-etal-2024-empirical}. To demonstrate the benchmark’s applicability, we evaluate four state‐of‐the‐art LLMs using a question‐answering task encompassing 3,300 questions. Our findings show that while the benchmarked LLMs can answer questions over event sets with a handful of events and explicit temporal references successfully, performance clearly deteriorates with larger event set length and when temporal references get less explicit. Notably, the \emph{vague} question category exhibits the lowest performance across all models.

The benchmark is publicly available at: \url{https://gitlab.ub.uni-bielefeld.de/s.kenneweg/TRAVELER}.
}




\maketitle

\section{Introduction}\label{sec1}
Reasoning about events is a fundamental aspect of natural language understanding \cite{vanLambalgen2006treatmentEvents}, as people often refer to past and future events pervasively in day-to-day communication. They refer to past and future events by different temporal references that need to be adequately resolved to identify a corresponding time point or event that the utterance makes reference to. 

Interpreting these references depends on the ability to address questions such as: 
\begin{enumerate}
    \item \label{Intro:Q1}Did Mary watch TV on the \textbf{13th of January 2023}?
    \item \label{Intro:Q2}When was the \textbf{last time} that Peter prepared a Risotto?
    \item \label{Intro:Q3}Who \textbf{just} prepared Risotto?
\end{enumerate}

The first question involves an expression, i.e. \emph{13th of January} that explicitly refers to a particular day and can be resolved without further disambiguation relative to another time point. 
The second question involves the expression \emph{last time} that needs to be resolved by taking into account all past events in which Peter did something, e.g. preparing a dish, and identifying the last event from a set of past events in which he cooked Risotto. 
Finally, the last question involves a vague temporal reference, the fuzzy temporal adverbial \emph{just} whose boundaries are not fully and unambiguously determined. While in case that somebody cooked Risotto 10 minutes ago it would be felicitous to utter that somebody `just' cooked Risotto, this would surely not be the case when the cooking took place several days ago. 

Such and other questions require to reason with respect to a chain or set of events that have happened in the past. We refer to this ability as \emph{Event Temporal Reasoning}\cite{wang2024trambenchmarkingtemporalreasoning}. Despite the importance of this ability for many applications, publicly available resources for systematically assessing the ability of automated systems to answer event-driven temporal questions remain scarce. We address this gap by introducing a synthetic benchmark comprising 3,300 English questions over sets of everyday household events. Our focus lies on two crucial dimensions. The first dimension is the degree of temporal explicitness: We classify questions by how explicitly they reference a point in time. Question~\ref{Intro:Q1} uses a maximally specific time stamp (i.e., a calendar date). In contrast, question~\ref{Intro:Q2} is more implicit, requiring the system to infer a specific date from the phrase \emph{`the last time'}, while question~\ref{Intro:Q3} uses the vague expression \emph{`just'}, introducing uncertainty into the reasoning process. The second important dimension is the size of the set of events: We systematically vary the length of the event sets from 5 to 100 events. Longer event sets challenge a system’s ability to maintain and use relevant information without a structured storage of past events.

Based on these two dimensions, we formulate the following hypotheses:

\begin{itemize}
\item H1: The performance of a system will degrade with decreasing level of explicitness of temporal references. 
\item H2: The performance of systems will be particularly low for questions involving vague temporal references. 
\item H3: The performance of a system will degrade for questions which require the consideration of longer event sets. 
\end{itemize}

We test these hypotheses using  Large Language Models (LLMs), which have recently shown impressive progress on reasoning tasks and temporal question answering \cite{wei2022emergent, saparov2023Language}. Concretely, we encode an event set into the model's prompt and then ask a question that probes its capacity to reason over that set. We compare four LLMs (\texttt{Gemma-7b-it}\cite{gemmateam2024gemma}, \texttt{Llama3-8B-Instruct}, \texttt{Llama3-70B-Instruct}\cite{Llama3website}, and \texttt{GPT-4-0125}\cite{openai2023gpt4}), examining how well they handle different lengths of event sets and varying degrees of temporal explicitness.

In summary our contributions are the following:
\begin{itemize}
    \item We tackle the problem of temporal reasoning over events involving temporal references of varying degrees of explicitness in the context of a Question Answering task
    \item We present a synthetic benchmark of 3,300 questions over common household events as a domain.
    \item We systematically analyze prompt engineering methods to find an effective prompt for the task. 
    \item We provide a comparative evaluation of four LLMs, reporting results for different event set lengths and levels of explicitness. 
\end{itemize}

Overall, our results corroborate our three hypotheses. We observe that the LLMs performance consistently degrades for larger event sets, suggesting that current models struggle with extended chains of reasoning in a single prompt. In addition, their performance declines noticeably when handling temporal references that are \emph{implicit}—and even more so when those references are \emph{vague}—compared to questions that include \emph{explicit} temporal references. These findings highlight the continuing need for targeted benchmarks and more advanced reasoning capabilities in LLMs to handle event-based temporal queries. 

This work is an extended version of our conference paper \cite{kenneweg_keod2024}.

\section{Related Work}
Events can be ontologically understood as occurrences in time involving participants who fulfill various roles, such as agent, patient, or beneficiary. In his foundational work, Davidson \cite{davidson2001logicalform} argued that action sentences should be interpreted as referring to events—entities that can be treated as objects to which participant roles are assigned.

Subsequent research has further explored the classification of event types and their internal structures. For instance, \citet{verbsandtimesvendler} introduced influential distinctions among event subtypes, including activities, achievements, and accomplishments. \citet{Moens1988TemporalOA} proposed a structured model of events comprising a preparatory phase, a culmination, and a consequent phase.


\subsection{Categories of Temporal Questions}

Temporal questions can be categorized in various ways depending on their inherent characteristics. According to \citet{sun2025timelinekgqacomprehensivequestionanswerpair}, one common way to distinguish them is based on their \textbf{Answer Type}, which can be either \emph{factual} or \emph{temporal}. Additionally, questions may involve specific types of \textbf{temporal relations}, such as Set Operations (union, intersection, negation), Allen's Temporal relations (relationships such as before or during), or Duration Operations (comparison or calculation of durations). 

Other works \cite{TempQuestions, saxena2021questionansweringtemporalknowledge} categorize temporal questions based on the nature of the \textbf{temporal reference}, which refers to how time is expressed or implied within a question. For instance, \citet{jia2024faithfultemporalquestionanswering} distinguishes between \textbf{explicit} and \textbf{implicit} temporal questions. Following previous work \cite{domaintemporaltagging, valueoftemporalinformation, domaintemporaltaggingforevent}, we make the following distinctions:

\begin{itemize}
    \item \textbf{Temporally explicit questions}, where the temporal expression refers unambiguously and independently to a specific point in time. For example: \emph{25th of December 2023}.
    
    \item \textbf{Temporally implicit questions}, where the temporal reference is context-dependent and requires additional knowledge for resolution. Examples include: \emph{Christmas 2023}, \emph{yesterday}, or \emph{Tom's birthday}.
\end{itemize}

Implicit temporal questions can be further divided into the following subcategories:

\begin{itemize}
    \item \textbf{Common sense knowledge} — questions where resolving the temporal reference (e.g., \emph{Christmas 2023}) requires general world knowledge.

    \item \textbf{Referential relative to speech time} — questions with expressions such as \emph{today}, \emph{yesterday}, or \emph{two days ago}, which depend on the time the question is spoken or written.
    
    \item \textbf{Referential relative to an arbitrary time point} — for example, \emph{two days before Christmas 2022}, which requires resolution relative to another temporal anchor.
    
    \item \textbf{Personal knowledge} — temporal expressions that require individual or private knowledge, e.g., \emph{Who watched TV on Tom's birthday?}
\end{itemize}

Furthermore, the temporal expression may vary in their degree of \textbf{vagueness}. As noted by \citet{TempQuestions}, some references such as \emph{``three weeks''} are concrete, whereas others like \emph{``several years''} or \emph{``some time ago''} are considered vague due to their fuzzy boundaries. 

In our benchmark, we include the question categories \textbf{explicit}, \textbf{implicit relative to speech time} and \textbf{vague}. 

\subsubsection{Vague Expressions}
Vagueness arises when there are borderline cases—situations in which ordinary speakers cannot definitively say whether an expression applies or does not apply. For example, the adjective \texttt{tall}, when used to describe people, is vague because there are individuals for whom it is neither clearly appropriate nor clearly inappropriate to describe them as tall.

While vagueness can occur across various parts of speech, much of the existing literature has focused on adjectives \cite{Kamp2016, Kennedy, Lassiter, Solt2012}. However, vagueness is also evident in other grammatical categories, such as quantifiers (e.g., many, few) and nouns (e.g., mountain).

This paper turns attention to a relatively underexplored area: vague temporal adverbials used in questions involving \textbf{vague} temporal references. Expressions like \emph{recently}, \emph{just}, \emph{some time ago}, and a \emph{long time ago} are vague, yet they have received limited attention in the broader literature on vagueness. 


A notable exception is the empirical study by \citet{kenneweg-etal-2024-empirical}. They investigated the applicability of vague temporal adverbials in English. For this purpose they have developed an online survey in which adult native English speakers were asked to rate the degree to which a certain fuzzy temporal adverbial could be applied to refer to an event that took place a certain amount of time ago. For instance, participants evaluated statements like: “Tom’s birthday was 1 day ago. Statement: Tom had his birthday recently.” Responses were recorded on a Likert scale, and the results illustrated the probability that a particular temporal adverbial would be used for a given event and elapsed time.

\subsection{Benchmarks for Event Temporal Reasoning}\label{subsec:temporalbenchmarks}
Temporal reasoning is the ability to understand, represent, and reason about time-specific data. One possibility to evaluate the ability of models to perform temporal reasoning is by question answering tasks involving temporal questions. A temporal question contains a temporal expression, temporal signal, or the answer is temporal \cite{Jia_2018}.

A number of benchmarks have been introduced to assess such capabilities. These include \emph{TimeQA}~\cite{chen2021datasetansweringtimesensitivequestions}, \emph{TempQuestions}~\cite{TempQuestions}, \emph{TimeQuestions}~\cite{TimeQuestions}, \emph{Test of Time}~\cite{fatemi2024test}, \emph{Date Understanding}, and \emph{Temporal Sequences}, the latter two are part of the \emph{Beyond the Imitation Game Benchmark (BIG-bench)}~\cite{bigbench}. 

The \emph{TRAM} benchmark~\cite{wang2024trambenchmarkingtemporalreasoning} offers a comprehensive suite by combining ten datasets to evaluate various aspects of temporal understanding in language, such as event order, duration, frequency, and temporal arithmetic. \citet{wang2024trambenchmarkingtemporalreasoning} categorize temporal reasoning into three levels: symbolic, commonsense, and event temporal reasoning. The focus of our work lies in the latter---event temporal reasoning---which specifically examines the relationships between events and their timing, as well as how events interact within the context of real-world scenarios. This type of reasoning typically demands the interpretation of a set of events.

Several existing benchmarks target event temporal reasoning. \emph{COMPLEXTEMPQA}~\cite{ComplexTempQA} presents over 100 million question-answer pairs centered on historical events. \emph{TRACIE}~\cite{zhou-etal-2021-temporal} challenges models to infer the temporal order of both explicit and implicit events in short narratives. \emph{COTEMPQA}~\cite{su2024livingmomentlargelanguage} evaluates models' ability to reason about events that occur simultaneously or overlap in duration, often involving complex temporal relationships. \emph{MenatQA}~\cite{wei-etal-2023-menatqa} is designed to test understanding across three dimensions: the duration of events (scope), their chronological sequence (order), and potential outcomes under alternative scenarios (counterfactuals).

\emph{MultiTQ}~\cite{chen-etal-2023-multi} explores reasoning across different temporal granularities---ranging from days to years---using knowledge graphs. \emph{TIQ}~\cite{tiqbenchmark} focuses on answering questions constrained by implicit temporal information, while \emph{TempQA-WD}~\cite{neelam2022benchmarkgeneralizableinterpretabletemporal} offers a benchmark over knowledge bases, grounded in Wikidata and supporting SPARQL queries with cross-knowledge-base evaluation. \emph{PATQA}~\cite{meem2024patquestionsselfupdatingbenchmarkpresentanchored} tests the handling of temporal questions that are either explicitly or implicitly tied to the moment of speech.

Despite this rich landscape of benchmarks, to the best of our knowledge, none provides a setting that evaluates an agent's ability to reason over extended sets of events involving different participants. Further, to the best of our knowledge, our benchmark is the only one that includes questions involving \emph{vague} temporal references.

\subsection{Large Language Models for Reasoning}
LLMs have demonstrated strong performance across a variety of reasoning tasks (see \citet{huang2023reasoning, plaat2024reasoninglargelanguagemodels} for recent surveys). These tasks range from symbolic operations—such as concatenating the final letters of words (e.g., Last Letter Concatenation\footnote{\url{https://huggingface.co/datasets/ChilleD/LastLetterConcat}})—to more complex forms of mathematical and arithmetic reasoning. Notable benchmarks in this domain include AQuA for algebraic problem solving \cite{aqua}, SVAMP for math word problems \cite{svamp}, and GSM8K for graduate-level word problems \cite{gsm8k}.

A consistent finding across studies is that reasoning performance tends to scale with model size \cite{wei2022emergent, saparov2023Language}. Additionally, techniques such as Chain-of-Thought prompting have been shown to significantly enhance LLM reasoning capabilities \cite{suzgun2022challenging}. Despite these advances, LLMs have not yet been systematically evaluated on tasks involving temporal reference resolution across extended event sets—a gap this paper aims to address.

However, LLMs continue to face challenges when dealing with reasoning tasks that closely mirror real-world scenarios, such as commonsense planning \cite{valmeekam2023planbench, joublin2023copal}. For instance, \citet{parmar2024logicbench} show that LLMs often fail to account for context in logical reasoning over natural language. Similarly, while they can handle tasks requiring single deductive steps, LLMs struggle with problems involving multi-step reasoning \cite{saparov2023Language}. This limitation becomes especially apparent in temporal reasoning contexts, as demonstrated by \citet{chu2024timebenchcomprehensiveevaluationtemporal}. Therefore, evaluating LLMs' ability to resolve both explicit and implicit temporal expressions—particularly in multi-event, multi-step reasoning scenarios—remains a critical and largely unexplored area of research, as underscored by recent work on temporal reasoning benchmarks \cite{chu2024timebenchcomprehensiveevaluationtemporal}.

\section{Benchmark Design}
We introduce \textbf{TRAVELER}, a temporal question answering benchmark designed to evaluate the event‐temporal reasoning capabilities of LLMs and other models. The benchmark comprises three categories of temporal questions, each with increasing levels of complexity: \textbf{explicit}, \textbf{implicit relative to speech time}, and \textbf{vague}. A key goal is to assess both how well a system resolves temporal references within each category and whether increasing the number of events in a scenario (i.e., the length of the event set) affects this performance. To this end, we generate event sets of varying lengths (Section~\ref{subsec:eventsets}) and pair each with corresponding temporal questions (Section~\ref{subsec:questions}). Each question is designed to fit into one of the three temporal question categories, enabling a systematic evaluation of temporal reasoning across different levels of complexity.

\subsection{Generation of Synthetic Event Sets}\label{subsec:eventsets}
To facilitate our experiments, we systematically generate synthetic events by randomly sampling from predefined sets of actions, agents, objects, locations, and timestamps. Our events are designed to reflect scenarios commonly occurring in a home environment, with each event instantiated as a tuple 
\(\langle \text{Event Type, Subject, Location, Timestamp} \rangle\). The types of events and potential fillers for the three roles subject and location are given in Table~\ref{tab:eventsparticipants}. Timestamps are provided as a Unix timestamp ranging from 2023-01-01 to 2023-09-29.

By iterating this process, we create sets of synthetic events of varying lengths (5, 10, 20, 30, 40, 50, 60, 70, 80, 90, and 100). As an illustrative example, one such event might be \(\langle \text{watch, film, mary, living room, 1695948843} \rangle\). Owing to the large sampling space --- especially for timestamps --- the likelihood of generating duplicate events remains negligible. 

\begin{table}[h]
\caption{Overview of used event types, agents and locations.} \label{tab:eventsparticipants}
\begin{tabular}{@{}l l|l|l@{}}
\toprule
\multicolumn{2}{c|}{\textbf{Event Type}} & \textbf{Subject} & \textbf{Location} \\
\midrule
        Watch & Film & \multirow{5}*{\shortstack[l]{Mary\\Tom\\Robot}} & \multirow{5}*{\shortstack[l]{Living Room\\Kitchen}} \\
        Eat & Risotto &  & \\
        Read & Book &  & \\
        Dance & Lively Salsa & &\\
        Store & Wine Bottle & &\\
        Drink & Juice &  & \\
        Chat with & Friend &  & \\
\botrule
\end{tabular}
\end{table}

\subsection{Question Generation}\label{subsec:questions}
We automatically generate questions for each event set, guided by the templates in Table~\ref{tab:Templates_QA}.

Our approach instantiates each of the four question templates in Table~\ref{tab:Templates_QA} with each temporal expression in Table~\ref{tab:temp_expr_categories}. Following the approach of \citet{chen-etal-2023-multi}, we cover multiple granularities of date representations for the temporal expressions of the \textbf{explicit} question category. For the \textbf{vague} category and the temporal expression \emph{just} the \emph{\textless temporal\_expression\textgreater} stands after \emph{\textless subject\textgreater}. Note that the fourth template (\emph{When was the last time \dots?}) is instantiated only for the category \textbf{implicit relative to speech time}. Consequently, we could generate 37 questions ($12*3+1=37$) per event instance over all three question categories.

For example, given the event:

\(\langle \text{Event Type: wash mug, Location: kitchen, Subject: tom, Timestamp: 1695852168} \rangle\),

we generate questions such as:
\begin{itemize}
  \item Who washed a mug in the kitchen on \emph{2023-08-16}?
  \item Who washed a mug in the kitchen \emph{in the year 2023}?
  \item When was the last time Tom washed a mug in the kitchen?
  \item Did Tom wash a mug in the kitchen \emph{yesterday}?
  \item Did Tom \emph{just} wash a mug in the kitchen?
  \item Did Tom wash a mug in the kitchen \emph{some time ago}?
  \item \dots
\end{itemize}


The ground-truth is derived automatically for each question (see Section~\ref{sec:gtgeneration} for details). In total, we generate 100 questions for each combination of event-set length and question category, resulting in $100*11*3=3300$ questions overallThe ground-truth is derived automatically for each question (see Section~\ref{sec:gtgeneration} for details). In total, we generate 100 questions for each combination of event-set length and question category, resulting in $100*11*3=3300$ questions overall.The ground-truth is derived automatically for each question (see Section~\ref{sec:gtgeneration} for details). In total, we generate 100 questions for each combination of event-set length and question category, resulting in $100*11*3=3300$ questions overall.

\begin{table}[h]
\caption{Templates for the Questions of the Question-Answer Set. The last template is only executed for the \emph{Implicit relative to speech time} question category (Reproduced from \citet{kenneweg_keod2024}).} \label{tab:Templates_QA}
\begin{tabular}{l|l}
\toprule
\textbf{Template} & \textbf{Return Type} \\
\midrule
        Who \textless event\_type\textgreater \textless location\textgreater \textless temporal\_expression\textgreater? & String \\
        Did \textless subject\textgreater \textless event\_type\textgreater \textless location\textgreater \textless temporal\_expression\textgreater? & Bool\\
        How often did \textless subject\textgreater \textless event\_type\textgreater  location\textgreater \textless temporal\_expression\textgreater? & Integer\\
        \midrule
        When was the last time \textless subject\textgreater \textless event\_type\textgreater \textless location\textgreater? & Date\\
\botrule
\end{tabular}
\end{table}

\begin{table}[h]
\caption{Temporal Expressions for the 3 categories of temporal questions. yyyy is the year with four digits, mm the month of the year with two digits, and dd the day of the month with two digits (Adapted from \citet{kenneweg_keod2024}).} \label{tab:temp_expr_categories}
\begin{tabular}{l|l}
\toprule
\textbf{Question Category} & \textbf{Temporal Expression}\\
\midrule
        \multirow{3}*{\shortstack[l]{\emph{Explicit}}} & on yyyy-mm-dd \\
                                & in yyyy-mm \\
                                & in the year yyyy \\
        \hline
        \multirow{5}*{\shortstack[l]{\emph{Implicit relative to speech time}}} & today \\
                                      & yesterday \\
                                      & this year \\
                                      & this month \\
                                      & last month \\
                \hline
        \multirow{4}*{\shortstack[l]{\emph{vague}}} 
                                      & just \\
                                      & recently \\
                                      & some time ago \\
                                      & long time ago \\
\botrule
\end{tabular}
\end{table}

\subsubsection{Ground Truth Generation and Accuracy Calculation}\label{sec:gtgeneration}

For the evaluation, we utilize two strategies. First, we assess how well LLMs identify events based on \emph{explicit} and \emph{implicit relative to speech time} temporal references, using a binary classification approach. Second, we evaluate \emph{vague} temporal expressions probabilistically, as no absolute true-or-false decisions can be made.

\paragraph{Explicit and Implicit Relative to Speech Time}
In order to evaluate whether the answer of a LLM is correct, for each question involving explicit and implicit (relative to speech time) temporal references, we automatically determine the correct answer by comparing the event type, subject, location, and event date of all events in the event set with the information specified in the question.

For example, given the question:

\begin{quote}
\textit{Did Tom eat Risotto in the living room on 2023-09-29?}
\end{quote}

the correct answer is \textit{yes} if there exists an event in the event set that matches the following criteria: \textless Subject:Tom, Event Type: eat risotto, Location: living room, Event Date: 2023-09-29\textgreater.

The LLMs response is evaluated using binary accuracy by comparing it's response with the correct answer.

\paragraph{Vague References}
For questions involving vague temporal references there is no definite ground truth however as the vague adverbials lack clear boundaries.  We opted thus for a probabilistic evaluation scheme that weights the answer of a system by the probability or degree to which the vague temporal adverbial applies to an event that took place a certain time ago. 

For this, following the methodology of \citet{kenneweg-etal-2024-empirical}, we conducted seven surveys—one for each event type (watching a film, eating risotto, reading a book, dancing salsa, storing a wine bottle, drinking juice, and chatting with a friend). Each survey involved 50 native speakers recruited via Prolific\footnote{https://www.prolific.com/}. They evaluated how appropriate each of the four vague temporal adverbials—\emph{just}, \emph{recently}, \emph{some time ago}, and \emph{long time ago}—are for describing events occurring at different time points in the past. Specifically, \emph{just} and \emph{recently} were tested with events occurring within the past two days, while \emph{some time ago} and \emph{long time ago} were assessed with events occurring up to one week prior. An example sentence from the survey, similar to the structure used by \citet{kenneweg-etal-2024-empirical}, was: \emph{Tom chatted with a friend in his living room \textless time\_span\textgreater ago. Statement: Tom chatted with a friend in his living room \textless adverbial\textgreater .}

\begin{figure}[h]
  \centering
  \includegraphics[scale=0.4]{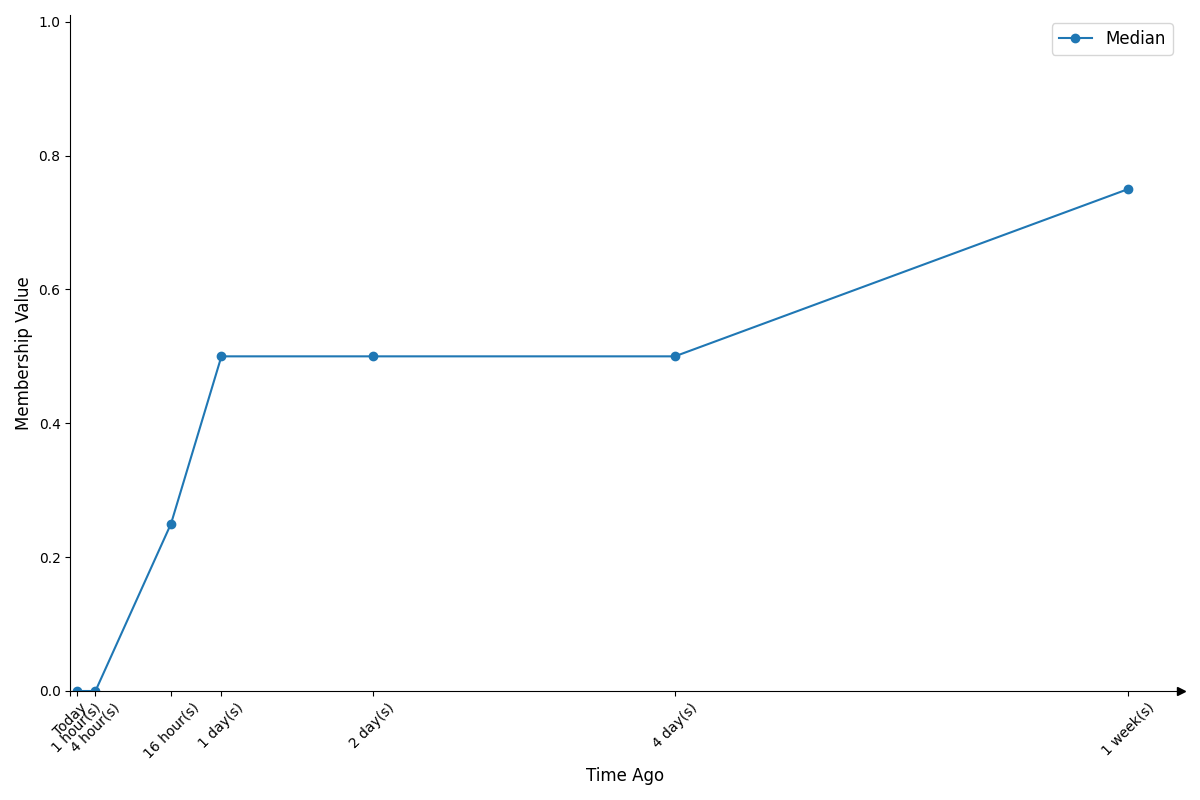}
  \caption{Median of the responses of all survey participants for the event type \emph{Eating Risotto} and adverbial \emph{long time ago}.}\label{fig.survey}
\end{figure} 

Figure~\ref{fig.survey} illustrates the probability distribution for the event type \emph{eating risotto} with the adverbial \emph{long time ago}. We take the median of all responses as the final probability value for whether a given adverbial describes an event that occurred a certain time in the past. 

Finally, we compute accuracy for each question type (e.g., \textit{Did \dots?}, \textit{Who \dots?}, \textit{How often \dots?}) using the following probabilistic approach:

Let \(G\) denote the \textbf{Ground Truth Set} containing all events \(e\) from the event set. Each event \(e\) has attributes \(\{S_e, T_e, L_e, D_e\}\), which are the  Subject \(S_e\) (e.g.\ ``Tom''), Event type \(T_e\) (e.g.\ ``eat risotto''), Location \(L_e\) (e.g.\ ``kitchen''), Date-time \(D_e\) (e.g.\ `2023-09-29 13:20'').

For a given subject $S$, event type $T$ and location $L$, we define the \textbf{subset} \(G' \subseteq G\) as follows:
\begin{equation}\label{EqSubset}
    G'(S,T,L) \;=\; \{\,e' \in G \;|\; S_{e'} = S,\ T_{e'} = T,\ L_{e'} = L \}.
\end{equation}

In order to determine the likelihood  \(p_{e,A}(\Delta t)\) that a certain temporal adverbial $A$ can be used to refer to an event of type $e$ that happened $\Delta t = D_{\text{ref\_time}} - D_e$ time units ago, we rely on the results of our surveys.


Given a subset \(G'\subseteq G\), we define the probability that \emph{none} of the events in \(G'\) is appropriately described by \(A\):
\begin{equation}\label{Eq:NoneOccurs}
\bar{P}_{G',A} \;=\; \prod_{e' \in G'} \bigl(1 - p_{e',A}(D_{\text{ref\_time}} - D_e)\bigr),
\end{equation}
Thus, the probability that \emph{at least one} event in \(G'\) fits \(A\) is described by the logical complement
\begin{equation}\label{Eq:OneOccurs}
P_{G',A}=1-\bar{P}_{G',A} \ .
\end{equation}

\paragraph{Did \dots?}
A question of the form: 
\[
\text{``Did \{S\} do \{T\} in the \{L\} \{A\}?''}
\]
is characterized by a specific subject \(S\), event type \(T\), location \(L\), and adverbial \(A\). Therefore we first compute the subset of events \(G'(S,T,L)\) that match \(S, T,\) and \(L\), as in \eqref{EqSubset}. 

When calculating the probability for this question, we calculate in the case of a LLMs "yes" response if the adverbial \(A\) applies to at least one event in \(G'\) and for a "no" response if it applies to none of the events in \(G'\). Accordingly using the probability from \eqref{Eq:OneOccurs}, that \emph{at least one} event in \(G'\) fits \(A\), we have:
\begin{eqnarray}\label{eq:didaccuracy}
P_{\text \tiny response}(\text{yes}\ |\ S,T,L,A)&=&P_{G',A} \\
P_{\text \tiny response}(\text{no}\ |\ S,T,L,A)&=&1-P_{G',A}
\end{eqnarray}
For the accuracy estimation of the LLM responses, we consequently use $P_{\text \tiny response}(\text{yes/no}\ |\ S,T,L,A)$.

Other types of questions out of the \emph{vague} question category are treated in analogous ways.

\subparagraph{\textbf{Example Did \dots?}} 


\noindent
In this example, we have this set \( G \) that contains all events mentioned in a prompt:
\[
G = \left\{
\begin{aligned}
   &\langle \text{S: Mary},\,
            \text{T: watch\_TV},\,
            \text{L: living\_room},\,
            \text{D: 2023\text{-}09\text{-}29\ 08{:}01} \rangle,\\
   &\langle \text{S: Tom},\,
            \text{T: eat\_risotto},\,
            \text{L: kitchen},\,
            \text{D: 2023\text{-}09\text{-}29\ 20{:}27} \rangle,\\
   &\langle \text{S: Ria},\,
            \text{T: read\_book},\,
            \text{L: living\_room},\,
            \text{D: 2023\text{-}06\text{-}11\ 12{:}44} \rangle,\\
   &\langle \text{S: Tom},\,
            \text{T: eat\_risotto},\,
            \text{L: kitchen},\,
            \text{D: 2023\text{-}09\text{-}22\ 22{:}27} \rangle,\\
   &\langle \text{S: Mary},\,
            \text{T: store\_wine\_bottle},\,
            \text{L: kitchen},\,
            \text{D: 2023\text{-}01\text{-}22\ 19{:}18} \rangle
\end{aligned}
\right\}
\]
To answer the question \emph{“Did Tom eat risotto in the kitchen a long time ago?”}, we extract a subset \( G' \subseteq G \) containing all events where the subject, location, and event type match the question. From the question, we identify the relevant components: \( S = \text{Tom} \), \( L = \text{kitchen} \), and \( T = \text{eat\_risotto} \). Thus, the filtered set becomes:

\[
G' = \left\{
\begin{aligned}
   &\langle \text{S: Tom},\,
            \text{T: eat\_risotto},\,
            \text{L: kitchen},\,
            \text{D: 2023\text{-}09\text{-}29\ 20{:}27} \rangle,\\
   &\langle \text{S: Tom},\,
            \text{T: eat\_risotto},\,
            \text{L: kitchen},\,
            \text{D: 2023\text{-}09\text{-}22\ 22{:}27} \rangle
\end{aligned}
\right\}
\]
Assuming a reference time of \( D_{\text{ref\_time}} = \text{2023-09-29 22:27} \), we compute the time difference \( \Delta t = D_{\text{ref\_time}} - D_e \) for each event \( e \in G' \), in order to estimate the probability that the adverbial \( A = \text{long\_time\_ago} \) applies. Using the likelihood \( p_{e,A}(\Delta t) \) from our carried out surveys (shown for \( T = \text{eat\_risotto} \) and \( A = \text{long\_time\_ago} \) in Figure~\ref{fig.survey}), we obtain:
\begin{align*}
&p_{e_1, A}(2 \text{ hours})= 0 \\
&p_{e_2, A}(1 \text{ week}) = 0.75
\end{align*}
If the LLM predicts \texttt{``yes''}, we calculate the combined probability (cf. Eq.~\eqref{eq:didaccuracy}):
\[
\begin{aligned}
P_{\text{\tiny response}}(&\text{yes} \mid S{=} \text{Tom},\, T{=} \text{eat\_risotto},\, L{=} \text{Kitchen},\, A{=} \text{long\_time\_ago}) \\
&= 1 - (1 - 0)\times(1 - 0.75) = 0.75
\end{aligned}
\]
Thus, the accuracy of this prediction is $0.75$.

\subsection{Prompting Strategies}\label{sec:promptingstrategies}
As a baseline prompting strategy, we employ a zero-shot prompt in which we specify only the expected answer of the LLM based on the question templates in Table~\ref{tab:Templates_QA}. The baseline prompt is provided in Figure~\ref{fig:zero-shotprompt}. Within this strategy, we systematically vary the level of temporal detail. Specifically, we distinguish two granularities: \emph{Date-Only} and \emph{Date-Extended}. 

In the \emph{Date-Only} condition, we present only the date (including hour and minute). In the \emph{Date-Extended} condition, we additionally provide the corresponding weekday and calendar week. For example:
\begin{quote}
\texttt{Date: 2023-08-11 10:57, Weekday: Friday, Calendar Week: 32}
\end{quote}

Besides altering the temporal granularity, we also vary how the events and their associated dates are presented to the model. In the \emph{Json} condition (see Figure~\ref{fig:zero-shotprompt}), each event is encoded in JSON format. In the \emph{Language} condition, the event and its temporal information are expressed in natural language. For the \emph{Date-Only} condition, this takes the form:
\begin{quote}
\texttt{On September 29, 2023 at 08:01, Mary watched a film in the living room.}
\end{quote}

Beyond this zero-shot prompting approach, we also explore an advanced prompting strategy based on Chain of Thought (CoT) reasoning. We define two CoT variants: \emph{CoT Review} and \emph{CoT Step-by-Step}. Under \emph{CoT Review}, the model is instructed to examine each event in the event set to identify any that match the query's event type, location, and date, subsequently returning the subjects of all matching events. For a ``Who\dots?'' query, this could be specified as:
\begin{quote}
\texttt{Review each event in the event set sequentially. If the event type, location, and date match the information in the question, record the subject of that event. At the end, return the subjects of all matching events.}
\end{quote}

In \emph{CoT Step-by-Step} reasoning, we extend the \emph{CoT Review} prompt by adding the instruction ``Let's think step by step,'' thereby encouraging a more explicit, stepwise reasoning process.

\begin{figure}[t]
  \centering
  \fcolorbox{owncolor!75}{owncolor!75}{
        \begin{minipage}{0.925\linewidth}
            \begin{small}
                 Today is the 2023-09-29 22:18. I will give you a list indicating events and when they have taken place (event set): \{Action: watch, Object: film, Location: living room, Subject: Mary, Date: 2023-09-29 08:01\}, \{Action: eat, Object: risotto, Location: kitchen, Subject: Tom, Date: 2023-09-28 14:27\}, \{Action: read, Object: book, Location: living room, Subject: Ria, Date: 2023-06-11 12:44\}, \{Action: dance, Object: lively salsa, Location: kitchen, Subject: Mary, Date: 2023-08-11 10:57\}, \{Action: store, Object: wine bottle, Location: living room, Subject: Tom, Date: 2023-09-01 20:44\}. Who watched a film in the living room on 2023-09-29? Answer with the the name of the subject or say 'nobody'.
            \end{small}
        \end{minipage}
    }
  \caption{Exemplary zero-shot prompt for an event set length of 5 events (Reproduced from \citet{kenneweg_keod2024}).}
  \label{fig:zero-shotprompt}
\end{figure}

\section{Experimental Setup \& Results}
\subsection{Experimental Plan \& Settings}
We evaluate four state-of-the-art LLMs: \texttt{Gemma-7b-it} \cite{gemmateam2024gemma}, \texttt{Llama3-8B-Instruct}, \texttt{Llama3-70B-Instruct} \cite{Llama3website}, and \texttt{GPT-4-0125} \cite{openai2023gpt4}. The experiments proceed in two phases. First, we assess all possible prompting strategies on \texttt{GPT-4} using event set lengths of 5 and 50 and the two question categories \emph{Explicit} and \emph{Implicit relative to speech time}. From this initial stage, we select the four most effective prompting strategies. In the second phase, these selected strategies are tested on each of the four models, with event set lengths ranging from 5 to 50 and the two question categories \emph{Explicit} and \emph{Implicit relative to speech time}, to identify the strategy that yields the best performance overall. Finally, using this top-performing strategy, we investigate how performance varies by question type, question category, and event set length. Furthermore, prompts are appropriately formatted for each model - for example, using system and user roles for \texttt{GPT-4} (\texttt{"role": "system", "content": "\dots"}, \texttt{"role": "user", "content": "\dots"}) and token-based turn delineations for \texttt{Gemma} (\texttt{\textless bos\textgreater \textless start\_of\_turn\textgreater user \dots \textless end\_of\_turn\textgreater \textless start\_of\_turn\textgreater model \dots})

\texttt{Llama3}\footnote{\url{https://huggingface.co/collections/meta-llama/meta-llama-3-66214712577ca38149ebb2b6}} and \texttt{Gemma}\footnote{\url{https://huggingface.co/google/gemma-7b-it}} experiments are performed on GPU, while \texttt{GPT-4} is accessed via its API. The \texttt{Llama3} model is evaluated in both its 8B and 70B instruction variants, and \texttt{Gemma} in its 7B instruction variant. \texttt{Llama3} and \texttt{Gemma} are obtained from HuggingFace without additional fine-tuning. We employ accuracy as our primary performance metric. To ensure reproducibility, each model is set to a temperature of 0 (or an equivalent deterministic setting). 

\subsection{Results}
We begin by analyzing the impact of all possible prompting strategies for \texttt{GPT-4} in Section~\ref{subsub:promptingstrategies}. Building upon these insights, Section~\ref{subsub:modelimpact} presents the results for all models using the four best-performing strategies identified in Section~\ref{subsub:promptingstrategies}. Next, Section~\ref{subsub:questioncategoriesresults} examines how question type influences the performance of the benchmarked LLMs. Finally, in Section~\ref{subsub:sequencelength}, we investigate the relationship between the length of the event set and model performance.

\subsubsection{Prompting Strategies}\label{subsub:promptingstrategies}
We consider a range of prompting strategies composed of three prompt types (\emph{zero-shot}, \emph{CoT Review}, \emph{CoT Step-by-Step}), two date encodings (\emph{Date-Only}, \emph{Date-extended}), and two event presentations (\emph{Json}, \emph{Language}), yielding 12 configurations in total. These configurations are evaluated using \texttt{GPT-4} with event set lengths of 5 and 50 for the two question categories \emph{Explicit} and \emph{Implicit relative to speech time}. Table~\ref{tab:accuracypromptengineering} presents the resulting accuracy scores. Across all prompting strategies, accuracy is higher with 5 events than with 50 events. Chain-of-thought prompts (\emph{CoT Review} and \emph{CoT Step-by-Step}) generally outperform the zero-shot baseline, whereas extended date encoding (\emph{Date-extended}) does not offer a clear benefit over simple date encoding (\emph{Date-Only}). The most effective configurations employ CoT-based prompting combined with \emph{Date-only} representations, irrespective of whether events are presented in \emph{Json} or \emph{Language} format. Figure~\ref{fig:final_prompt} shows the final prompt with these prompting strategies for an event set length of 5 events. 

\begin{figure}[h]
  \centering
  \fcolorbox{owncolor!75}{owncolor!75}{
        \begin{minipage}{0.925\linewidth}
            \begin{small}
            Review each event out of the event set sequentially. If the action, object, location and date of an event match the information in the question, record the subject of that event. At the end return the subjects of all matched events. Today's date is September 29, 2023, and the time is 22:18. I have a list of events (event set) that have occurred in the past, including who did what, where and when: On September 29, 2023 at 08:01, Mary watched a film in the living room. On September 28, 2023 at 14:27, Tom ate a risotto in the kitchen. On June 11, 2023 at 12:44, Ria read a book in the living room. On August 11, 2023 at 10:57, Mary danced a lively salsa in the kitchen. On September 01, 2023 at 20:44, Tom stored a wine bottle in the living room. Now, I want to know: Who watched a film in the living room on September 29, 2023?
            \end{small}
        \end{minipage}
    }
  \caption{Exemplary final prompt for an event set length of 5 events, the question template "Who\dots ?" and the question category \emph{Explicit} (Reproduced from \citet{kenneweg_keod2024}).}
  \label{fig:final_prompt}
\end{figure}

\begin{table}[h]
\caption{Accuracy of all possible prompts for \texttt{GPT-4-0.125} averaged for the two question categories \emph{Explicit} and \emph{Implicit relative to speech time} over event set lengths of 5 and 50 events. The last column is the average of the accuracy for 5 and 50 Events. The 4 highest results are marked in bold (Reproduced from \citet{kenneweg_keod2024}).} \label{tab:accuracypromptengineering}
\begin{tabular}{l l l|l l|l}
\toprule
         \multirow{2}{*}{\shortstack[l]{Prompting\\ Strategy}} & \multirow{2}{*}{\shortstack[l]{Date\\ Information}} & \multirow{2}{*}{\shortstack[l]{Event\\ Presentation}} & \multicolumn{2}{|c|}{Events} & \multirow{2}{*}{Average} \\
         & & & 5 & 50 & \\
\midrule
        Zero-Shot & Date-Only & Json & .97 & .67 & .82\\
        Zero-Shot & Date-Only & Language & .96 & .67 & .82\\
        Zero-Shot & Date-Extended & Json & .97 & .64 & .81\\
        Zero-Shot & Date-Extended & Language & .96 & .68 & .82\\
        \textbf{CoT Review} & \textbf{Date-Only} & \textbf{Json} & .97 & .71 & \textbf{.84}\\
        \textbf{CoT Review} & \textbf{Date-Only} & \textbf{Language} & .94 & .71 & \textbf{.83}\\
        CoT Review & Date-Extended & Json & .95 & .68 & .82\\
        CoT Review & Date-Extended & Language & .93 & .71 & .82\\
        \textbf{CoT Step-by-Step} & \textbf{Date-Only} & \textbf{Json} & .94 & .71 & \textbf{.83}\\
        \textbf{CoT Step-by-Step} & \textbf{Date-Only} & \textbf{Language} & .95 & .71 & \textbf{.83}\\
        CoT Step-by-Step & Date-Extended & Json & .94 & .66 & .80\\
        CoT Step-by-Step & Date-Extended & Language & .94 & .70 & .82\\
\botrule
\end{tabular}
\end{table}

\subsubsection{Model Impact}\label{subsub:modelimpact}
Table~\ref{tab:promptengineering4bestallmodels} reports accuracy scores for the four best prompting strategies across all models for event sets of 5 and 50 and the two question categories \emph{Explicit} and \emph{Implicit relative to speech time}. 
Notably, the larger models (\texttt{GPT-4} and \texttt{Llama3-70B}) achieve the highest accuracy (83\%--84\% for \texttt{GPT-4}; 84\%--90\% for \texttt{Llama3-70B}), with \texttt{Llama3-70B} showing a slight advantage over \texttt{GPT-4}. 
In contrast, \texttt{Gemma} and \texttt{Llama3-8B} exhibit lower overall performance, with accuracies ranging from 63\% to 86\% (\texttt{Gemma}) and 68\% to 74\% (\texttt{Llama3-8B}). 
Based on these findings, we select the configuration that yields the highest average performance across all models for subsequent experiments: \emph{CoT Review}, \emph{Date-Only}, and \emph{Language} encoding.
\begin{table}[h]
\caption{Accuracy of the 4 best performing prompt configurations for \texttt{GPT-4-0.125} on all evaluated LLMs and the two question categories \emph{Explicit} and \emph{Implicit relative to speech time} averaged over event set lengths of 5 and 50 events. The highest result for each model and the highest average result is marked in bold (Reproduced from \citet{kenneweg_keod2024}).}
    \label{tab:promptengineering4bestallmodels}
\begin{tabular}{l l l|l l l l | l}
\toprule
         \multirow{2}{*}{\shortstack[l]{Prompting\\ Strategy}} 
         & \multirow{2}{*}{\shortstack[l]{Date\\ Information}} 
         & \multirow{2}{*}{\shortstack[l]{Event\\ Pres.}} 
         & \multirow{2}{*}{\shortstack[l]{\texttt{Gemma}\\ \texttt{-7b-it}}} 
         & \multirow{2}{*}{\shortstack[l]{\texttt{Llama3}\\ \texttt{-8B-Instr.}}} 
         & \multirow{2}{*}{\shortstack[l]{\texttt{Llama3}\\ \texttt{-70B-Instr.}}} 
         & \multirow{2}{*}{\shortstack[l]{\texttt{GPT-4}\\ \texttt{-0125}}}
         & \multirow{2}{*}{\shortstack[l]{Ave-\\ rage}} \\
         & & & & & & & \\
\midrule
        CoT Rev. & Date-Only & Json & \textbf{.68} & .68 & .86 & \textbf{.84} & .76\\
        \textbf{CoT Rev.} & \textbf{Date-Only} & \textbf{Lang.} & \textbf{.68} & \textbf{.74} & .88 & .83 & \textbf{.78}\\
        CoT Step. & Date-Only & Json & .63 & .69 & .84 & .83 & .75\\
        CoT Step. & Date-Only & Lang. & .65 & .72 & \textbf{.90} & .83 & .77\\
\botrule
\end{tabular}
\end{table}

\subsubsection{Type of Questions}\label{subsub:questioncategoriesresults}
Table~\ref{tab:accuracyquetionreturn} summarizes the results for the three question categories, 
\emph{Explicit}, \emph{Implicit relative to speech time} and \emph{vague}, as well as for the  individual question templates listed in Table~\ref{tab:Templates_QA}. 

\paragraph{Impact of degree of explicitness.}
The accuracy for \emph{explicit} questions ranges from 75\% (\texttt{Llama3-8B}) to 92\% (\texttt{Llama3-70B}). By contrast, questions involving expressions that must be interpreted relative to the speech time exhibit a marked drop in performance, with accuracy varying between 34\% (\texttt{Gemma}) and 74\% (\texttt{Llama3-70B}). The lowest performance occurs in the \emph{vague} category, ranging between 26\% (\texttt{Gemma}) and 45\% (\texttt{Llama3-70B}). Shifting from explicit to implicit temporal references reduces model accuracy by approximately 40\%.

\paragraph{Results by template type.}
Regarding question templates, the \emph{Did \dots?} template produces the highest accuracy, ranging from 68\% (\texttt{Gemma}) to 86\% (\texttt{GPT-4}). 
The lowest scores are 34\% (\texttt{Gemma}, \emph{When was the last time \dots?}), 46\% (\texttt{Llama3-8B}, \emph{Who \dots?}), 47\% (\texttt{GPT-4}, \emph{Who \dots?}) and 65\% (\texttt{Llama3-70B} \emph{Who \dots?} and \emph{How often did \dots?}).

\begin{table}[h]
\caption{Accuracy for the different question categories and question templates averaged over all evaluated event set lengths. The right column is the average of all models. The highest results of each model for each question category, question template and vague adverbial are marked in bold (Adapted from \citet{kenneweg_keod2024}).} \label{tab:accuracyquetionreturn}
\begin{tabular}{l|l l l l l | l}
\toprule
         & &  \multirow{2}{*}{\shortstack[l]{\texttt{Gemma}\\ \texttt{-7b-it}}} & \multirow{2}{*}{\shortstack[l]{\texttt{Llama3-8B}\\ \texttt{-Instruct}}} & \multirow{2}{*}{\shortstack[l]{\texttt{Llama3-70B}\\ \texttt{-Instruct}}} & 
         \multirow{2}{*}{\shortstack[l]{\texttt{GPT-4}\\ \texttt{-0125}}} &
         \multirow{2}{*}{\shortstack[l]{Ave-\\ rage}} \\
         & & & & & \\
\midrule
        \multirow{3}{*}{\shortstack[l]{Question\\ Category}} &
        Explicit & \textbf{.84} & \textbf{.75} & \textbf{.92} & \textbf{.84} & .84 \\
        & Implicit relative\dots & {.34} & {.58} & {.74} & .64 & {.58} \\
        & Vague & {.26} & {.45} & {.41} & .42 & {.39} \\
\midrule
         \multirow{4}{*}{\shortstack[l]{Question\\ Templates}} &
           Who \dots? & .44 & .46 & .65 & .47 & .51 \\
         & Did \dots? & \textbf{.68} & \textbf{.77} & \textbf{.82} & \textbf{.86} & .78\\
         & How often did \dots? & .35 & .52 & .65 & .57 & .52\\
         & When was the \dots? & .34 & .53 & .66 & .75 & .57\\
\botrule
\end{tabular}
\end{table}

\subsubsection{Event Set Length}\label{subsub:sequencelength}
\begin{figure}[h]
  \centering
  \includegraphics[scale=0.45]{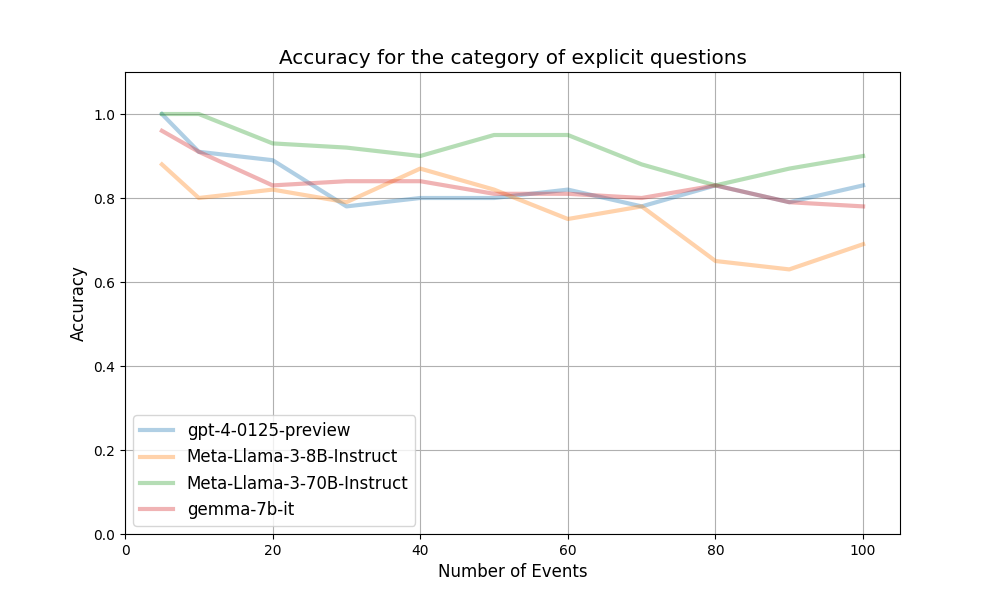}
  \caption{Accuracy for the \emph{Explicit} question category depending on set length (Reproduced from \citet{kenneweg_keod2024}).}\label{fig.accuracyovereventhistoryabsolute}
\end{figure}

\begin{figure}[h]
  \centering
  \includegraphics[scale=0.45]{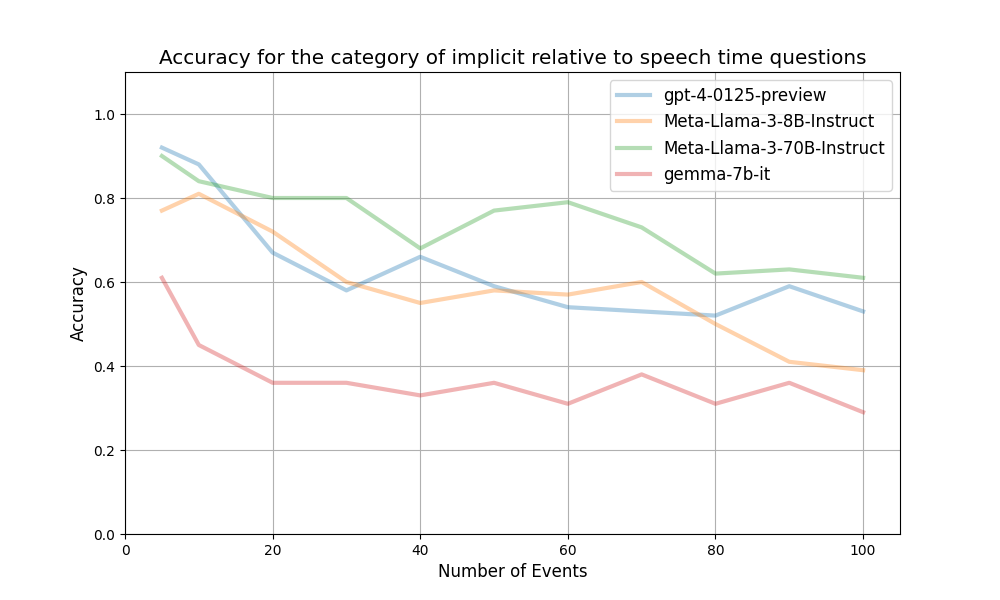}
  \caption{Accuracy for the \emph{Implicit relative to speech time} question category depending on set length (Reproduced from \citet{kenneweg_keod2024}).}\label{fig.accuracyovereventhistorynow}
\end{figure}

\begin{figure}[h]
  \centering
  \includegraphics[scale=0.45]{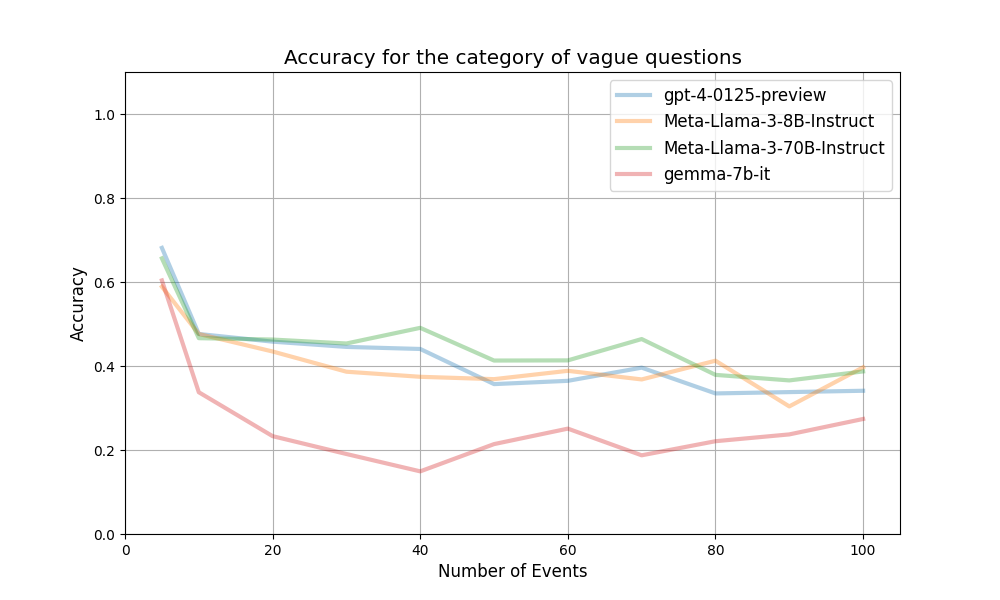}
  \caption{Accuracy for the \emph{vague} question category depending on set length.}\label{fig.accuracyovereventhistoryvague}
\end{figure}

Figures~\ref{fig.accuracyovereventhistoryabsolute}, ~\ref{fig.accuracyovereventhistorynow} and and~\ref{fig.accuracyovereventhistoryvague} show the results for all evaluated question categories, (\emph{Explicit}, \emph{Implicit relative to speech time} and \emph{vague}), across all evaluated event set lengths (5, 10, 20, 30, 40, 50, 60, 70, 80, 90, 100). We can see that model performance decreases substantially with increasing set length. 

For the \textbf{Explicit} questions, the reduction in accuracy from 5 to 100 events ranges 
from 18\% (\texttt{Gemma}) to 10\% (\texttt{Llama3-70B}). This corresponds to a stepwise decrease 
of approximately 1.0\% (\texttt{Llama3-70B}) to 1.8\% (\texttt{Gemma}) per increment in set length.

In the \textbf{Implicit relative to speech time} category, the performance drop is even more 
pronounced, varying from 39\% (\texttt{GPT-4} and \texttt{Llama3-8B}) to 29\% (\texttt{Llama3-70B}). 
On a stepwise basis, this corresponds to a decrease between 2.9\% (\texttt{Llama3-70B}) and 
3.9\% (\texttt{GPT-4} and \texttt{Llama3-8B}) as the number of events considered increases from 5 to 100.

In the \textbf{vague} category, the performance drop ranges between 19\% (\texttt{Llama3-8B}) and 34\% (\texttt{GPT-4}). On a stepwise basis, this corresponds to a decrease between 1.9\% (\texttt{Llama3-B}) and 3.4\% (\texttt{GPT-4}) as the number of events considered increases from 5 to 100.

\section{Discussion}\label{sec:discussion}

Our findings strongly support our three hypotheses outlined in this work. 
Regarding the first hypothesis (Impact of temporal explicitness), we observe that for all models, questions containing \emph{implicit relative to speech time} references lead to a 26\% decrease in performance compared to \emph{explicit} references. When the reference is \emph{vague}, the performance drops by around 45\%, underscoring the difficulty LLMs face in interpreting ambiguous temporal expressions. By contrast, explicit temporal references seemingly provide more direct cues, allowing the models to align the question date precisely with the event record. The fact that the performance for \emph{vague} references further declines by 19\% relative to other implicit references confirms our second hypothesis and underlines a specific challenge in resolving inherently ambiguous expressions.
Regarding the last hypothesis (Impact of event set size), we see that as the size of the event set grows, model performance generally decreases. This trend is expected, given that LLMs lack a mechanism to store or access prior events via explicit memory. The degree of this performance degradation varies by model; for instance, \texttt{GPT-4} and \texttt{Llama3-8B} show the most pronounced drop of 39\% in the \emph{Implicit relative to speech time} category when moving from 5 to 100 events. The observed local performance improvements in certain specific set lengths should be studied in future work.

Considering the different prompting strategies, although performance variations remain within 6\%, we consistently find that \emph{chain-of-thought} (\emph{CoT}) prompting \cite{wei2023chainofthought,suzgun2022challenging} yields better results. Moreover, a \emph{Date-Only} format for event dates outperforms \emph{Date-Extended}, possibly because the additional temporal details (e.g., weekday) remain unused in our questions and simply expand the prompt. Finally, presenting events in natural language (versus JSON) appears more effective, likely reflecting the LLMs’ training bias toward conventional text formats rather than structured data.

Interestingly, questions using the \emph{Did\dots?} template achieve the highest accuracy across all models, presumably due to their binary, yes/no nature. By contrast, tasks requiring more extensive reasoning—such as enumerating participants (\emph{Who\dots?}) or counting occurrences (\emph{How often did\dots?})—exhibit about a 30\% lower success rate. Questions of type \emph{When was the last time\dots?} demand identifying and comparing all relevant event instances, which also proves to be a major hurdle.

Finally, we see that the largest models typically achieve the best results, reaffirming the importance of parameter scale. Notably, \texttt{Llama3-70B} slightly outperforms \texttt{GPT-4} in the \emph{explicit} and \emph{implicit relative to speech time} question category, despite having fewer parameters (70\,B vs.\ 1760\,B). This finding suggests that beyond a certain threshold, factors other than sheer model size may become critical for success. Additional research is needed to elucidate what specific model characteristics or training regimens make \texttt{Llama3-70B} so good at event-based temporal reasoning. For the \emph{vague} question category \texttt{Llama3-8B} surprisingly has a slightly better (3-4\%) accuracy than \texttt{Llama3-70B} and \texttt{GPT-4}. This could be an indication that in the case of ambiguous expressions the number of parameters is not so important as for the other question categories.

\section{Conclusion \& Future Work}\label{sec:conclusion}
We have presented \textbf{TRAVELER} a benchmark for assessing the ability of systems to perform event-based temporal reasoning in a question answering setting. We investigated the performance of four state-of-the-art LLMs on our benchmark. Our results confirm our three hypotheses: while LLMs perform well on questions with \emph{explicit} temporal expressions, they struggle with \emph{implicit} references and have the most problems with \emph{vague} references. Furthermore, increasing the size of the event set significantly declines performance.

In light of these findings, several directions for future work emerge. One possibility is to enhance LLMs with an explicit memory component that can store and access event information more reliably. Approaches like that of \citet{xiong2024large}, where they construct a temporal graph from historical event data to enable explicit memory retrieval and chain-of-thought reasoning, offer a promising avenue. Another complementary strategy is to integrate formal temporal reasoning modules or function calls into LLM workflows.

Moreover, a temporal training framework for sequence-to-sequence language models has been proposed by \citet{tan-etal-2023-towards}. They combine a pre-training phase with a temporal span extraction task—encouraging the model to focus on key temporal and entity spans—followed by task-specific fine-tuning and subsequent time-sensitive reinforcement learning. This multi-step process could further enhance LLMs' capacity to handle event temporal reasoning tasks.

Also, instead of manually performing prompt engineering like we have done, using automated tools like DSPy \cite{khattab2023dspycompilingdeclarativelanguage} is another fruitful research direction. Such tools can systematically generate and refine prompts, potentially enabling more reliable performance gains. 

Finally, the benchmark should be extended to include the other question categories (\emph{Questions requiring common sense knowledge}, \emph{Referential relative to an arbitrary time point}, \emph{Personal Knowledge}) and the performance of LLMs for them should be evaluated.

\backmatter

\backmatter

\bmhead{Funding}
This study was funded by the Honda Research Institute Europe. 
\bmhead{Code availability}
The code is publicly available in the GitHub repository mentioned in the abstract.
\section*{Declarations}
\bmhead{Ethical approval}
Approval for the performed surveys was obtained from the ethics committee of the Bielefeld University. The procedures used in this study adhere to the tenets of the Declaration of Helsinki.
\bmhead{Competing Interests}
The authors have no competing interests to declare that are relevant to the content of this article.

\bibliography{sn-bibliography}

\end{document}